\documentclass[10pt,twocolumn,letterpaper]{article}

\usepackage{cvpr}
\usepackage{times}
\usepackage{epsfig}
\usepackage{graphicx}
\usepackage{amsmath}
\usepackage{amssymb}
\usepackage{subcaption}
\usepackage{pgfplots, tikz}
\usepackage{booktabs}
\usepackage{multirow}
\usepackage[normalem]{ulem}
\usepackage{enumitem}

\usepackage[normalem]{ulem}
\usepackage{soul}
\usepackage{xstring}

\sethlcolor{magenta}
\makeatletter
\def\SOUL@hlpreamble{%
    \setul{\dp\strutbox}{\dimexpr\ht\strutbox+\dp\strutbox\relax}%
    \let\SOUL@stcolor\SOUL@hlcolor
    \SOUL@stpreamble
}
\makeatother

\newcommand{\addref}[1]{\textcolor{magenta}{ [REF] }}

\usepackage[pagebackref=true,breaklinks=true,letterpaper=true,colorlinks,bookmarks=false]{hyperref}

\cvprfinalcopy 

\begin{document}

\title{Rotation Equivariant Siamese Networks for Tracking}

\author{Deepak K. Gupta\footnotemark[1], Devanshu Arya,\footnotemark[2]  and Efstratios Gavves\footnotemark[1]\\
$^{*}$QUVA Lab, University of Amsterdam, The Netherlands\\
$^{\dagger}$Informatics Institute, University of Amsterdam, The Netherlands\\
{\tt\small \{d.k.gupta,d.arya,e.gavves\}@uva.nl}
}

\maketitle

\begin{abstract}
    Rotation is among the long prevailing, yet still unresolved, hard challenges encountered in visual object tracking. The existing deep learning-based tracking algorithms use regular CNNs that are inherently translation equivariant, but not designed to tackle rotations. In this paper, we first demonstrate that in the presence of rotation instances in videos, the performance of existing trackers is severely affected. To circumvent the adverse effect of rotations, we present rotation-equivariant Siamese networks (RE-SiamNets), built through the use of group-equivariant convolutional layers comprising steerable filters. SiamNets allow estimating the change in orientation of the object in an unsupervised manner, thereby facilitating its use in relative 2D pose estimation as well. We further show that this change in orientation can be used to impose an additional motion constraint in Siamese tracking through imposing restriction on the change in orientation between two consecutive frames. For benchmarking, we present Rotation Tracking Benchmark (RTB), a dataset comprising a set of videos with rotation instances. Through experiments on two popular Siamese architectures, we show that RE-SiamNets handle the problem of rotation very well and outperform their regular counterparts. Further, RE-SiamNets can accurately estimate the relative change in pose of the target in an unsupervised fashion, namely the in-plane rotation the target has sustained with respect to the reference frame. Code and data will be made available at \url{https://github.com/dkgupta90/re-siamnet}. 
\end{abstract}

\section{Introduction}
The task of visual object tracking with Siamese networks ~\cite{bertinetto2016fully,tao2016siamese}, also referred as Siamese tracking, transforms the problem of tracking into similarity estimation between a template frame and sampled regions from a candidate frame.
Siamese trackers have recently gained popularity in the field of visual object tracking, especially because of their strong discriminative power obtained from similarity matching. This is the primary reason most of the state-of-the-art trackers are based on this framework \cite{bertinetto2016fully, gupta2020tackling, li2019siamrpn++, li2018high,  tao2016siamese}.

Although Siamese trackers are generally shown to work well, they are still prone to failure under challenges such as partial occlusion \cite{kuipers2020hard}, scale change \cite{sosnovik2020scale} or when one of the two inputs is rotated. 

This paper focuses on handling the adverse affects of in-plane rotation of objects on the performance of Siamese trackers.
Object rotation is considered to be amongst the hardest challenges of tracking with no effective solution took. date. It can commonly occur in real-life scenarios, especially when the camera records from the top, as in drones, where either the object is rotating or the camera itself. Egocentric videos are another example, where large head rotations can cause the target to rotate. 


\begin{figure}
    \centering
    \includegraphics[scale=0.3]{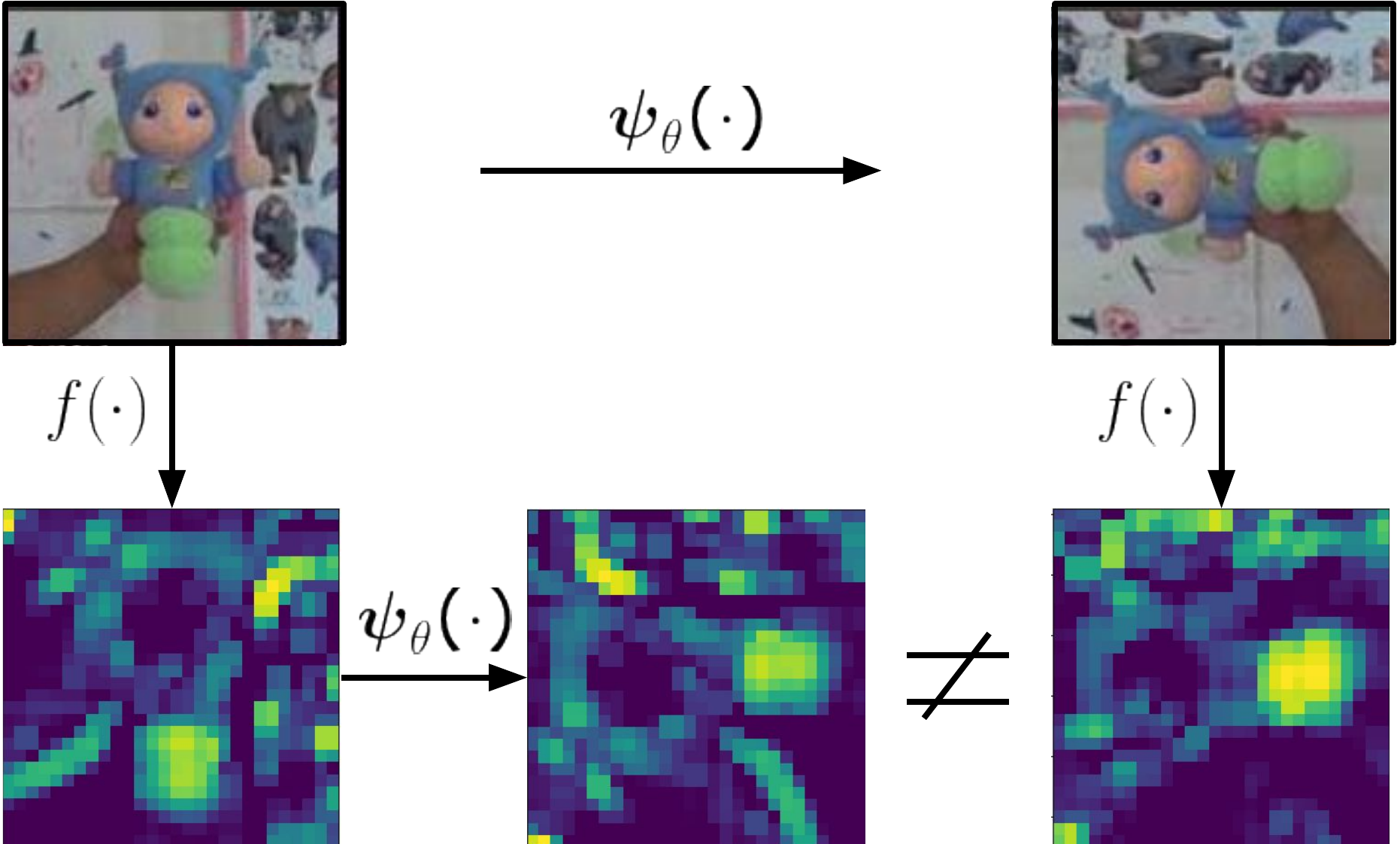}
    \caption{Example demonstrating rotation non-equivariance in regular CNN models used in object tracking, $\psi_{\theta}(f(\cdot)) \neq f(\psi_{\theta}(\cdot))$. Here $f(\cdot)$ and $\psi_{\theta}(\cdot)$ denote the neural network encoding function and rotation transform, respectively.}
    \label{fig1}
\end{figure}

The CNN architectures used in Siamese trackers are not inherently equivariant to in-plane rotations of the target. The implication is that the model may perform well on object orientations that are represented in the training set, but may fail on other previously unseen orientations. This happens because the latent encoding obtained from the network for such cases might not be representative of the input image itself. Example demonstrating this issue is shown in Figure \ref{fig1}. Further, even if it were equivariant, the cross-correlation step in traditional Siamese trackers would still fail to perform an accurate matching between the template and candidate images due to rotational shift between them. 

A straightforward approach to enforce learning of rotated variants is to use training datasets where in-plane rotations occur naturally or through data augmentation. However, as highlighted in \cite{laptev2016ti}, there are several limitations of data augmentation. First, such procedures would require learning separate representations for different rotated variants of the data. Second, the more variations are considered, the more flexible tracker model needs to be to capture them all. This means a significant increase in training data and computational budget. Further, such an approach  would make the model invariant to rotations, thus making the predictions unreliable when the target is surrounded by similar objects, \emph{e.g.}, tracking a fish in a school of fishes. \\
This paper aims at incorporating the property of rotation equivariance in the existing Siamese trackers. This built-in feature would then allow the trackers
to capture the rotation variations from the start itself without the need of additional data augmentation. Rotation equivariant networks have been studied widely in the context of image classification \cite{cohen2016group,cohen2016steerable,weiler2019general,weiler2018learning,worrall2017harmonic}. Drawing inspiration from these works, we introduce rotation equivariance for the task of object localization in videos. We exploit the concept of group-equivariant CNNs \cite{cohen2016group}, and use steerable \mbox{filters \cite{weiler2018learning}} to make the Siamese trackers equivariant to rotations. This way of incorporating rotation equivariance induces built-in sharing of weights among the different groups of rotations and adds an internal notion of rotation in the model (referred further as \emph{RE-SiamNet}). 

Interpreting the template image as the static memory of the tracking model, RE-SiamNets know beforehand how the encoding should be represented for a discrete set of rotations. In the absence of other challenges such as illumination variation and occlusion, the target appearance would match exactly at one of the discrete rotations, and is expected to contain only small errors for other intermediate angles. This property increases the discriminative power of the trackers towards differences in orientation (in-plane rotation) of the target. Beyond this,
RE-SiamNet can be used for relative 2D pose estimation of  objects in videos, interchangeably also referred in this paper as relative orientation estimation of objects. RE-SiamNets are equivariant to translations and rotations, and these properties combined with the structure of Siamese networks allow capturing the change in pose of the target in 2D. Further, we propose an additional motion constraint on the rotational degree of freedom and demonstrate that it allows to obtain better temporal correspondence in videos. 

It is important to note that most current datasets, especially in tracking, contain very limited to no instances of rotation. Thus, for benchmarking the performance of models in presence of in-plane rotations, we present Rotating Object Benchmark (ROB), a set of videos focusing on in-plane rotations. Annotations include bounding boxes of the target object as well as its orientation in every frame.
To further summarize, the contributions of this paper are:

\begin{itemize}[noitemsep]
    \item We give a brief introduction to equivariant convolutions networks. We then extend the theory to obtain rotation-equivariant Siamese architectures (RE-SiamNets) that feature in-plane rotation equivariance.
    \item We show that RE-SiamNets estimate the relative change in 2D pose of any rotating object in a unsupervised manner. Further, we introduce an additional motion constraint to improve temporal correspondence in videos.
    \item For benchmarking, we present Rotating Object Benchmark (ROB), a novel dataset comprising sequences with significant in-plane rotations of the target.
    \item Through incorporating in two existing Siamese tracking methods, we show that rotation equivariance can provide significant improvements in tracking performance and accurately estimate the orientation changes.
\end{itemize}

\section{Related Work}

\textbf{Siamese tracking. }Object tracking aims at estimating the trajectory of an arbitrary  target in a video given only its initial state in a video frame \cite{kristan2015visual}. Most of the recent object tracking algorithms use Siamese networks and track the object based on similarity matching \cite{dong2019quadruplet, fan2019siamese, guo2017learning, shen2019visual, wang2019spm, wang2019fast, zhang2018structured}. Such algorithms estimate a general similarity function between the feature representations learned for the target template and the candidate search region in a given frame. 

 The first Siamese trackers, SINT \cite{tao2016siamese} and SiamFC \cite{bertinetto2016fully}, used twin subnetworks with shared parameters and calculated dot product similarities between the feature maps of the template and the candidate frame. Held et al. \cite{held2016learning} introduced a detection-based Siamese tracker in which the similarity function was modeled as a fully-connected network. They applied extensive data augmentation  for learning a generalized function for multiple object transformations. Valmadre et al. \cite{valmadre2017end},  introduced CFNet which expanded SiamFC using a differentiable correlation filter layer. 
All of these trackers were able to get good performance in terms of object deformation compared with the trackers without online updating, but were not suitable in fast tracking
situations. Some of the subsequent methods such as \cite{he2018twofold, li2018high, wang2018learning, zhu2018distractor} discarded online updating, and turned to learn a robust feature representation instead. This allowed the aforementioned methods to perform high speed tracking using Siamese networks.

\textbf{Challenges of tracking. } There are several challenges encountered in visual object tracking that can affect the performance of the designed tracking algorithms. A detailed study highlighting some of the most important challenges was presented in \cite{smeulders2013visual}. These include illumination variation, in-plane and out-of-plane rotations of the target, occlusion, clutter and confusion due to several similar objects, among others. With recent large-scale training datasets such as LaSOT \cite{fan2019lasot} and TrackingNet \cite{muller2018trackingnet}, and state-of-the-art deep learning trackers, several of these challenges can be addressed up to a high degree of accuracy. For example, trackers such as SiamRPN++ \cite{li2019siamrpn++} and DiMP \cite{bhat2019learning} exhibit strong discriminative power with the use of deep CNN backbones, and have been found to tackle most of the challenges. However, some challenges such as occlusion and target rotation still remain to be solved. Recent works related to tackling occlusion tracking are \cite{gupta2020tackling} and \cite{kuipers2020hard}. In this paper, we focus on the challenge of target rotation.

 \textbf{Equivariant CNNs. } 
Recently, several works have tried to directly incorporate equivariance into the network’s architecture to capture various transformations. In this paper, we focus on rotation-equivariant CNNs which has gained popularity in image classification \cite{dieleman2016exploiting,cohen2016steerable}, texture classification \cite{marcos2016learning}, boundary detection \cite{worrall2017harmonic} and image segmentation \cite{laptev2016ti}. Dieleman et al. \cite{dieleman2016exploiting} included 4 operations  into existing networks to enrich both the batch- and feature dimension with their transformed versions. Cohen et al. \cite{cohen2016group} firstly introduced group-convolutional layers where feature maps resulting from transformed filters were treated as functions of the corresponding symmetry-group. However, in this method the computational cost was directly proportional to the group size, and this issue was resolved with steerable filters \cite{cohen2016steerable, weiler2018learning}. A detailed study providing a general theory of equivariance across various existing methods is provided in \cite{weiler2019general}. In this paper, we study rotation equivariance in the context of object tracking.

In real-life scenarios, tracking a target object is very challenging, especially since it can undergo transformations beyond translation, such as in-plane and out-of-plane rotations, occlusion and scale change. Unless the network has an internal mechanism to handle these transformations, the template matching similarity can degrade significantly in a Siamese network. Recent Siamese trackers \cite{li2019siamrpn++, zhang2019deeper} have implicitly or explicitly focused on making the trackers translational equivariant, i.e. a translation of the input image must result in the proportional translation of the corresponding feature space.  The importance of translation equivariance is to reduce the positional bias during training, so that location of the target is easier to recover from the feature space. SiamRPN++ \cite{li2019siamrpn++} proposed a training strategy which removes the spatial bias introduced in non fully-convolutional backbones. Further, \cite{zhang2019deeper} showed that existing tracking models induce positional bias, which breaks strict translation equivariance. Sosnovik et al. \cite{sosnovik2020scale}, introduced scale-equivariant Siamese trackers which is crucial when the camera zooms its lens or when the target moves into depth. 
We argue that in-plane rotations is also an important challenge in tracking, especially when the videos are recorded using drone cameras, other videos recorded from top view, cameras mounted on rotating objects and egocentric videos. To the best of our knowledge, rotation equivariance in the context of tracking has never been studied, and we address it in this paper.



\begin{figure*}
\centering
\includegraphics[scale=0.42]{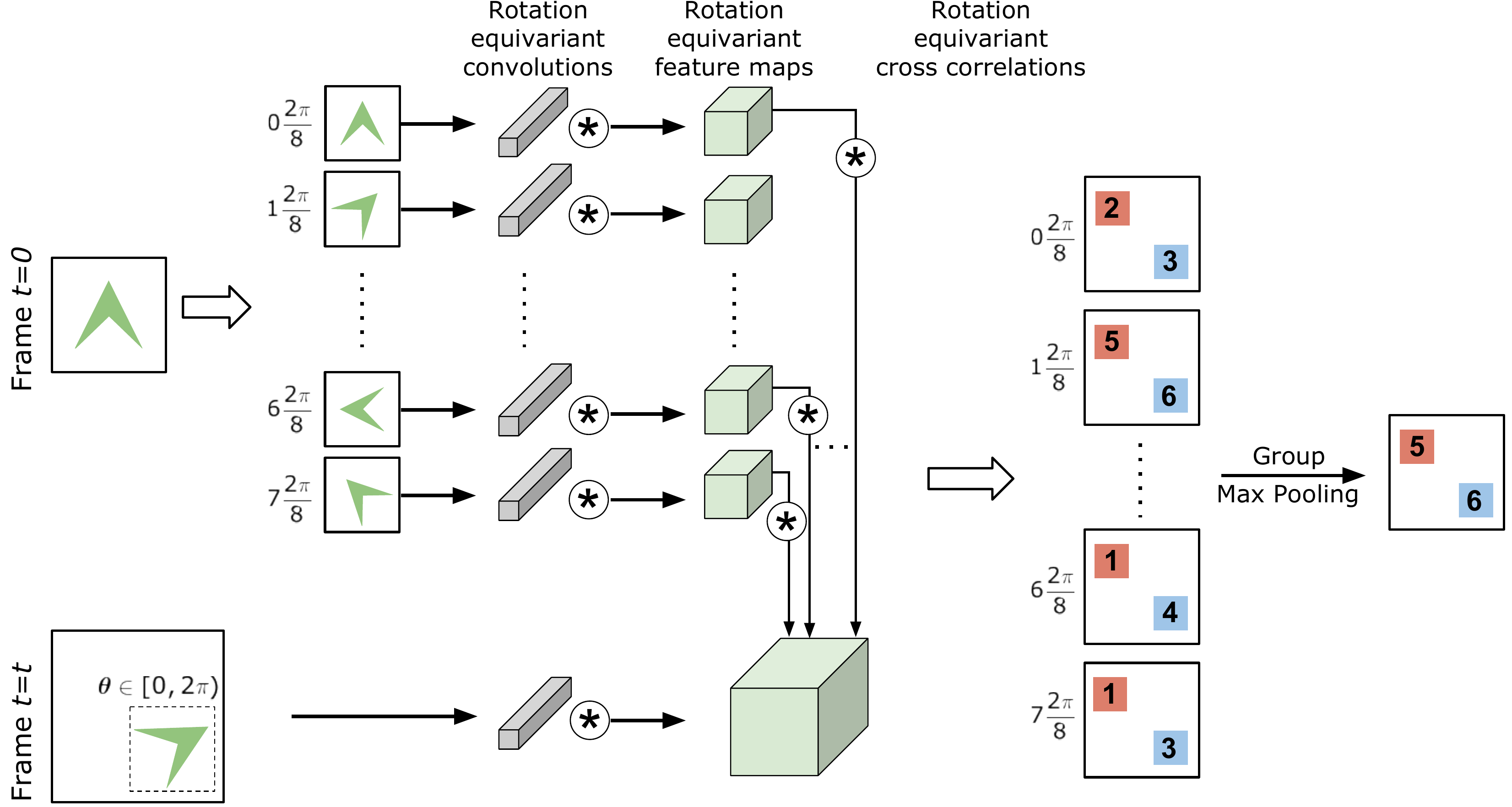}
\caption{Schematic representation of RE-SiamNet typically designed for object tracking. On the template head, multiple equidistant rotated variants of the original template image are used.}
\label{img_resiam}
\end{figure*}

\section{Rotation Equivariant CNNs}
We first provide some basic background knowledge on equivariance and rotation equivariance in CNNs required to formulate our tracker. For a more general overview we refer the interested reader to \cite{weiler2018learning}.

\textbf{Equivariance.}The property of equivariance requires functions to commute with the actions of a symmetry group acting on its domain and codomain. For any given transformation group $G$, a mapping function $f:X\xrightarrow{} Y$ is equivariant if it satisfies
\begin{equation}
    f(\psi_g^Y(x)) = \psi_g^Y(f(x)) \enskip g \in G, x \in X,
\end{equation}
where $\psi^{(\cdot)}_g$ denotes a group action in the respective space. For invariance, $\psi^{(\cdot)}_g$ will be an identity mapping.

For clarity, we take translation equivariance as an example. In this example, $f$ stands for the convolutional neural network function and $\psi_g$ denotes the translation group. Example actions from this group include for example, moving one pixel left, or one towards right, or an action comprising shift of several pixels. In this manner, an infinite number of actions can be defined within the translation group. Making the network equivariant to translations
leads to reduced sample complexity and facilitates generalization of the model against translational variations. 

It is important to note that there are several other transformations beyond translation that can be built in the model to improve robustness, if the effects of these transformations are present in the data and the task. Examples include rotations, reflections and scale change. For generalization over any of these transformations, equivariance needs to be enforced on the respective transformation group.
In this work we focus on rotation equivariance.

\textbf{Rotation equivariance.}
One of the more robust ways of enforcing rotation equivariance in CNNs is through the use of steerable filters \cite{weiler2018learning}.
Steerable filter CNNs (SFCNNs) extend the notion of weight sharing from the translation group to rotations as well.
For rotation equivariance with steerable filters, the network must perform convolutions with different rotated versions of each filter.
In this case weight sharing helps the model to generalize better.

Steerable filters not only facilitate efficiently computing responses for an arbitrary number of discrete filter rotations $\Lambda$, but they also exhibit strong expressive power as well.
A filter $\Psi$ is rotationally steerable if its rotation by an arbitrary angle $\theta$ can be expressed in terms of a fixed set of atomic functions~\cite{freeman1991design, weiler2018learning}. In our network, we employ circular harmonics $\psi_{jk}$ defined as
\begin{equation}
\psi_{jk}(r, \phi) = \tau_j(r)e^{\text{i}k\phi},
\end{equation}
where $\phi \in (-\pi, \pi]$ and $j=1,2,\hdots, J$ allows to control the radial part of the basis functions.
Further, the $(r, \phi)$ refers to transformed version of ($x_1, x_2$) in polar coordinates and $k \in \mathbb{Z}$ denotes the angular frequency.
The benefit of circular harmonics is that now we can simply express rotations on $\psi_{jk}$ as a multiplication with a complex exponential,
\begin{equation}
    \rho_{\theta}\psi_{jk}(x) = e^{-\text{i}k\theta}\psi_{jk}(x).
\end{equation} 
Note that for clarity purpose, we express $\psi_{jk}(\cdot)$ as $\psi_{jk}(x)$.

Each learnt filter is then constructed as a linear combination of the elementary filters, 
\begin{equation}
    \Psi(x) = \sum_{j=1}^J \sum_{k=0}^K w_{jk}\psi_{jk}(x),
\end{equation}
with weights $w_{jk} \in \mathbb{C}$. For rotation by $\theta$, the composed filter can be steered through phase manipulation of the elementary filters,
\begin{equation}
    \rho_{\theta}\Psi(x) = \sum_{j=1}^J \sum_{k=0}^K w_{jk}e^{-\text{i}k\theta}\psi_{jk}(x).
    \label{eq_filtering}
\end{equation}
A single orientation of the filter can be obtained by taking real part of $\Psi$, denoted as $\text{Re} \Psi(x)$.

\section{Rotation Equivariant Siamese Trackers}

\subsection{Proposed Formulation}

For trackers that rely on similarity matching with Siamese networks, the resultant heatmap $h(z, x)$ is
\begin{equation}
h(z, x) = f(z) * f(x),
\end{equation}
where $z$ and $x$ denote the template image and the candidate frames, respectively, $f(\cdot)$ is the encoding function of the Siamese network, and $*$ denotes the convolution operation. 

Figure \ref{img_resiam} presents the schematic representation of our RE-SiamNet framework for object tracking.
Architecturally, we start from and modify the basic SiamFC \cite{bertinetto2016fully} model due to its simple design.
The basic SiamFC comprises the following modular layers: input, convolutional layers, and a cross-correlation of the outputs from the two Siamese heads.
For our rotational Siamese tracker, we replace these layers with rotation equivariant modules. Further, we introduce a group max pooling module that selects the cross-correlation encoding for the most appropriate orientations among the multiple heatmaps generated in our setup. Details related to these modules follow below.

\textbf{Rotation equivariant input. } The candidate head of the network takes a single search image as input. However, the template head is modified to not just take one template image as an input, rather a set of its $\Lambda$ rotated variants defined by the set $Z$, where $Z = \{z_1, z_2, \hdots, z_{\Lambda}\}$. 
Instead of taking all possible rotation versions $Z$ of the template target, we could also first compute the feature $f(z)$ of the original target, then rotate $f(z)$. In theory, this is supported by rotation equivariant networks. In practice, however, the spatial resolution of $f(z)$ is very low, typically $6 \times 6$ or $7 \times 7$ pixels. As a result, there will be artifacts at the corners and edges because of the crudeness of the transformation. Instead, it yields more accurate feature maps if, when creating $Z$ in the first frame, we first rotate the whole frame (not just the target) centering about the target, and then crop. Since this is only performed on the target branch, it can be pre-computed during the inference phase. 

Each input image $I$, as stated above, comprises $C$ channels, where each channel is represented as $I_c$ and $c \in \{1, 2, \hdots, C\}$. This input is then convolved with $\hat{C}$ rotated filters $\rho_\theta \Psi_{\hat{c}c}^{(1)}$, where $\hat{c} \in \{1, 2, \hdots, \hat{C}\}$. Based on Eq. \ref{eq_filtering}, the resultant features obtained before applying nonlinear activation will be
\begin{align}
    y_{\tilde{c}}^{(1)}(x, \theta) = \text{Re}\sum_{c=1}^C \sum_{j=1}^J \sum_{k=0}^{K_j}w_{\hat{c}cjk} e^{-\text{i}k\theta}(I_c * \psi_{jk} )(x),
    \label{eq_filtering_input}
\end{align}
where the filters are then rotated variants at equidistant orientations $\theta$ represented by the set $\Theta = \{0, \Lambda, \hdots, 2\pi\frac{\Lambda - 1}{\Lambda}\}$. The bias term $\beta^{(1)}_{\hat{c}}$ and nonlinearity $\sigma$ are then applied to obtain the feature map at the first layer $\zeta_{\hat{c}}^{(1)}$.
%

\textbf{Rotation equivariant convolutions.} Feature maps resulting from Eq. \ref{eq_filtering_input} are processed further using group convolutions, generalizing spatial convolutions over a wider set of transformation groups. Similar to the first layer, steerable filters are defined on the group as 
\begin{align}
    y_{\hat{c}}^{(l)}(x, \theta) = & \text{Re}\sum_{c=1}^C \sum_{\phi \in \Theta} \sum_{j,k}w_{\hat{c}cjk,\theta-\phi} e^{-\text{i}k\theta}( \nonumber \\
    & \zeta_c^{(l-1)}(\cdot, \phi) * \psi_{jk} )(x).
    \label{eq_gcnn1}
\end{align}
The additional index $\theta-\phi$ introduced in Eq. \ref{eq_gcnn1} for the weight tensor facilitates the group convolution operation along the rotation dimension. It involves transforming the functions on the group through rotating them spatially. 

\textbf{Rotation equivariant pooling. }The output of the last group convolutional layer is further processed through pooling over the rotation dimension. Unlike the  conventional classification tasks, pooling is not performed along the spatial dimension to preserve the rotation equivariance.

\textbf{Rotation equivariant cross-correlation. }From the two subnetworks of the Re-SiamNet module, we obtain two sets of feature maps, $\{\phi(z)\}$ and $\phi(x)$, where $\{\phi(z)\}$ is the set containing feature maps at $\Lambda$ orientations. Next, $\{\phi(z)\}$ and $\phi(x)$ are convolved to obtain $\{\hat{h}(z,x)\}$, a set of $\Lambda$ heatmaps, where $h_i(z,x) = \phi(z_i) * \phi(x)$. Next, $\{\hat{h}(z,x)\}$ is processed with a global maxpooling operation to obtain the final output heatmap $h(Z, x)$. The global maxpooling operation identifies the maximum value in $\{\hat{h}(z,x)\}$ and selects the feature map that contains it. 

\vspace{0.5em}
By introducing the aforementioned modules, we obtain the rotation equivariant Siamese tracker.
Again, we emphasize that the tracker is equivariant to \emph{in-plane rotations}, as out-of-plane rotations require knowledge of the 3D scene to be integrated in the network. Next, we describe the training and inference of rotation equivariant Siamese trackers.

\subsection{Constructing RE-SiamNet Framework}
We outline below the steps to design RE-SiamNet framework using the rotation equivariant modules described in the earlier section. 
\begin{enumerate}[noitemsep]
\item Identify the precision of the tracker in terms of discriminating between different orientations of the rotational degree of freedom. We consider here $\Lambda$ rotation groups, based on which RE-SiamNets would be perfectly equivariant to angles defined by the set $\Theta = \{(i-1)\cdot 360/\Lambda\}^{\Lambda}_{i=1}$. 

\item Define the non-parametric encoding $\phi(\cdot)$ based on existing Siamese trackers. Based on the choice of $\phi(\cdot)$, the discriminative power of trackers varies. 

\item Replace all the convolutional layers of $\phi(\cdot)$ with the rotation-equivariant modules \footnote{For implementing rotation-equivariant modules in this paper, we use the \texttt{e2CNN} Pytorch library \cite{weiler2019general} available at \texttt{https://github.com/QUVA-Lab/e2cnn}}.

\item Instead of a single convolution to generate $h(z,x)$, $\Lambda$ convolutions are performed to generate $\Lambda$ different heatmaps. 

\item Perform Global max-pooling over the  feature maps to generate $h(Z,x)$, which is then processed to localize the target.
\end{enumerate}

Note that depending on the choice of the tracker head, processing operation on $h(Z, x)$ can differ. For example, for trackers such as SINT \cite{tao2016siamese} and SiamFC \cite{bertinetto2016fully}, target instance from the previous frame is fitted at different scales and aspect ratios, and the best among them is chosen. For other trackers such as SiamRPN \cite{li2018high} and SiamRPN++ \cite{li2019siamrpn++}, a region-proposal module is added that regresses the bounding box using a neural network head. In our tracking architecture, rotation equivariance needs to be only maintained up to $h(Z, x)$, thus it can work with any of these methods.

\section{Unsupervised Relative Rotation Estimation}
\textbf{Unsupervised 2D pose estimation. }The inherent design of RE-SiamNets allows to obtain an estimate of the relative change of 2D pose of the target in a fully unsupervised manner. This information can be obtained from the result of the group maxpooling step. Let $i \in \{1, 2, \hdots, \Gamma\}$ denote one of $\Lambda$ orientations of the template image. Then, $i$ is the number of rotation groups by which the pose of the template differs from that of its appearance in the candidate image, if:
\begin{equation}
h(Z, x) = \hat{h}(z_i, x) = \text{group-maxpool}(\{h(z, x)\}).
\end{equation}
The difference in pose expressed in terms of rotational angle $\theta_{\text{diff}}$ is then $i \cdot 360/\Gamma$.
Assuming that the actual in-plane rotation of the target is $\theta_c$, the error in prediction in degrees is bounded as $|\theta_{\text{diff}}-\theta_c| \leq \frac{360}{2\Lambda}$. Thus, for larger values of $\Lambda$, error in the estimation of pose decreases.

\textbf{Rotational Motion Consistency. }An important advantage is that RE-SiamNets provide a novel motion constraint that can be used to improve temporal correspondence in object tracking. To reiterate, Siamese trackers are mostly based on similarity matching with only weak temporal correspondence introduced through localizing the search area in any candidate frame around the target location in the previous frame and penalizing the changes in translation and scale between two consecutive frames. With RE-SiamNets, we explore the applicability in improving the temporal consistency through imposing restrictions on the rotational motion. This is achieved during the selection of $\theta_{\text{opt}} \in \Theta$ among the $\Lambda$ orientations. Let $\theta_{t, \text{opt}} = \theta_{t,i}$, where $\theta_{t,i}$ refers to the $i^\text{th}$ orientation in frame $t$. For the next frame, rather than selecting $\theta_{t+1, \text{opt}}$ from the full set $\Theta$, a constraint can be imposed such that $\theta_{t+1, \text{opt}} \in \{ \theta_i \}$. Index $i$ here is constrained to the set $\{i_{t,\text{opt}}-\gamma,\hdots, i_{t,\text{opt}}-1, i_{t,\text{opt}}, i_{t,\text{opt}}+1, \hdots, i_{t,\text{opt}}+\gamma\}$ such that $\gamma$ is the maximum change in number of orientations allowed in either directions between two consecutive frames. This constraint ensures that the orientation does not change by more than $\gamma$ groups between two successive frames.

\section{Rotating Objects Benchmark (ROB)}
\begin{figure*}
\centering
\includegraphics[scale=0.45]{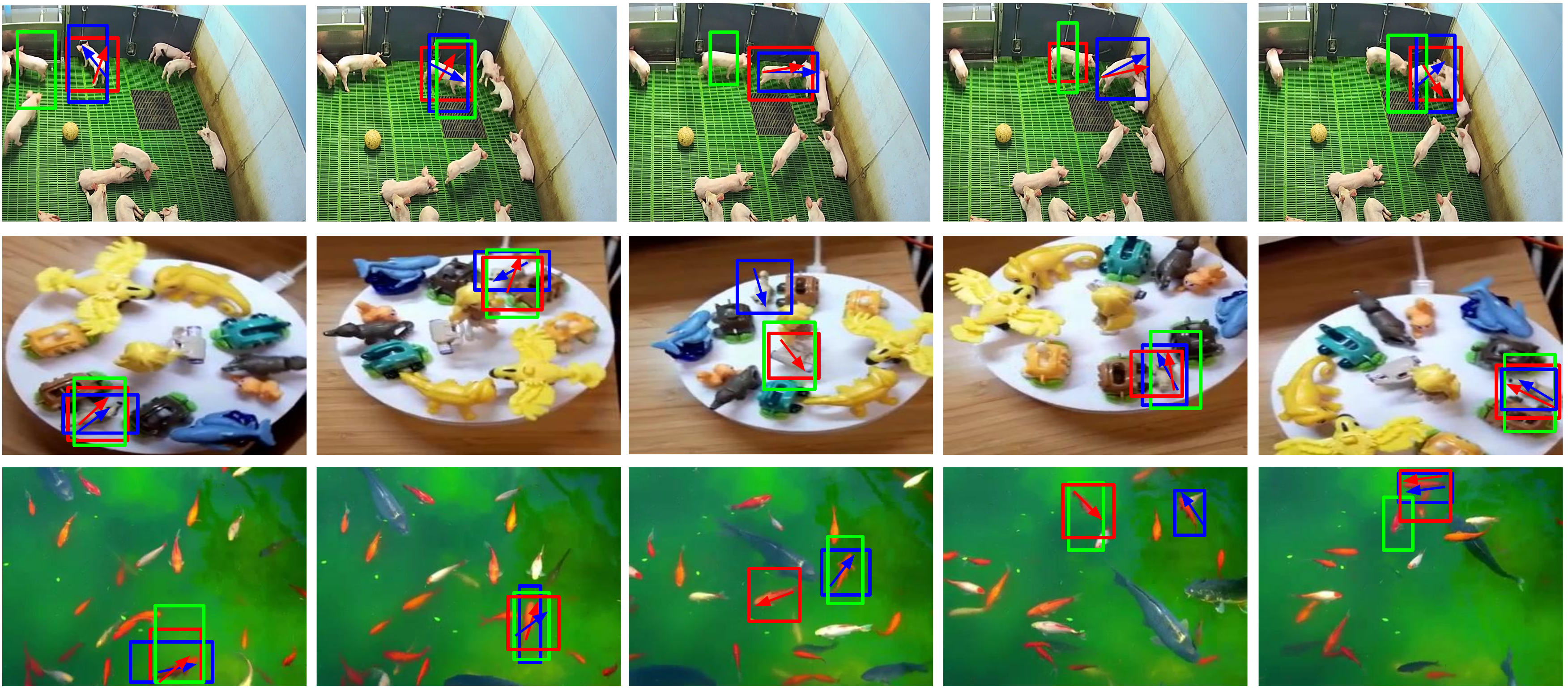}
\caption{Example frames from 3 sequences of ROB dataset showing the ground truth bounding box (blue), and predictions obtained using SiamFC\protect\cite{bertinetto2016fully}(green) and RE-SiamFC using 8 rotation groups (red). Further, blue and red arrows show the ground truth pose estimate and the prediction obtained using RE-SiamFC, respectively.}
\label{fig_rob_results}
\end{figure*}
State-of-the-art  benchmarks mostly do not contain rotation annotations. To evaluate RE-SiamNets as well as to enable future benchmarking on rotation sensitive tracking, unsupervised rotation estimation and rotation stabilization. We present Rotating Objects Benchmark (ROB) consisting of real world video sequences with large-scale variations in in-plane rotation of target objects.

ROB dataset is a collection of short video clips comprising multiple objects in diverse scenarios, where the target object undergoes rotation due to a rotating camera or/and an in-plane rotation of the object itself. In each video, the camera moves around the objects, capturing its different angles of rotation. The dataset consists of 35 video sequences with over 10,000 annotated frames and 15 object categories, ranging from a wide range of real-world scenarios such as livestock monitoring, cycling and aeroplanes. 

Sequences from ROB dataset are densely annotated in a semi-automated manner, with each frame providing objects location using bounding box coordinates, as well as information about its orientation with respect to the frame.
%
%
To annotate orientation change, a one-head arrow is drawn along one of the axes of target in the first frame, and consistently followed in rest of of the frames. This allows to compute the orientation change between the appearances of the target in any two frames of the sequences.


\section{Experiments}

We validate rotation equivariant Siamese trackers in tracking and estimation of relative 2D orientation changes. 
We first compare with the non-rotation equivariant version of the trackers, including SiamFC and SiamFCv2 \cite{bertinetto2016fully} and SiamRPN++ \cite{li2019siamrpn++}.
The proposed design philosophy, however, is general and any Siamese tracker can benefit.
Moreover, we compare with DiMP \cite{bhat2019learning} that attains SOTA results on standard tracking benchmarks. 

\textbf{Training. } 
All rotation equivariant variants of SiamFC are trained on the GOT-10k \cite{huang2019got} training set.
To train SiamRPN++, we trained a rotation equivariant version of ResNet50 architecture on ImageNet.
The SiamRPN++ model was then trained using this backbone on sets of COCO \cite{Lin2014eccv}, ImageNet DET \cite{Russakovsky2015ijcv}, ImageNet VID and YouTube-BoundingBoxes Dataset \cite{Real2017cvpr}

\textbf{Evaluation.}
To evaluate how well the proposed RE-SiamNets perform in presence of frequent in-plane rotations, we test them on ROB, Rot-OTB100 and Rot-MNIST datasets. Rot-OTB100 dataset is built by rotating each frame of OTB100 videos by 0.5 degree with respect to its previous frame. Rot-MNIST involves superposition of 3-5 MNIST digits on GOT-10k image backgrounds, and the digits translate and rotate randomly but in a smooth manner. Details related to the generation of these two datasets, as well as results on ROT-MNIST are provided in the supplementary section of this paper. To demonstrate that adding RE-SiamNets do not degrade the performance of trackers with respect to other challenges, we test them on tracking benchmarks that include OTB100 \cite{wu2013online} and GOT-10k \cite{huang2019got}. 



\textbf{Implementation Details}
%
To design RE-SiamNets, we adapt the existing models by replacing the regular CNN layers with rotation equivariant layers and using a group-pooling layer to output features at single orientation for every input. These rotation equivariant modules are added using the $\texttt{e2cnn}$ pytorch library \cite{weiler2019general}. For base Siamese trackers, we use SiamFC \cite{bertinetto2016fully}, its variant SiamFCv2, and SiamRPN++ \cite{li2019siamrpn++}. Here and henceforth, we use the prefix `RE-' to refer to the rotation equivariant version of a tracker.

%
For most experiments presented in this paper, we use RE-SiamFC. The base tracker SiamFCv2 differs from SiamFC in terms of the filter sizes and the number of convolutional layers. The former comprises only 4 convolutional layers with filter sizes of 9, 7, 7 and 6. The reason behind choosing this variant is to experiment with models involving largers filters, since these are known to work well for rotation equivariant CNNs \cite{weiler2019general}. Full details on the architecture of SiamFC and SiamFCv2 are provided in th supplementary section of this paper. We further point out that unless specifically differentiated, we will occasionally refer SiamFC and SiamFCv2 under the same name of SiamFC. 
We experiment with rotation groups of $\Lambda=4, 8, 16$ for SiamFC and $\Lambda=4$ for SiamRPN++. 

All RE-SiamNet implementations described in this paper are trained using stochastic gradient descent method. The methods follow the same training configurations as those of their base trackers. Exceptions include training of RE-SiamFC with R16 for 150 epochs using batch size of 16. Further, the rotation equivariant ResNet50 backbone was trained on ImageNet for only 50 epochs due to limited computational time. All models were trained on machines equipped with either 1 or 4 GPU Titan X GPUs.
Details on optimization can be found in the supplementary material.


\subsection{Rotation Equivariance in Tracking}

\begin{table}
\small
	\begin{center}
		\begin{tabular}{ l l  cc } 
			\toprule
			Model & Type & Succ & Pr \\ 
			\midrule
			\multirow{4}{*}{SiamFC \cite{bertinetto2016fully}} &  - & 0.315 &  0.523  \\
			& R4 & 0.360  & 0.629 \\
			& R8 & 0.423  & 0.676   \\
			\midrule
			\multirow{4}{*}{SiamFCv2} & - & 0.288  & 0.473 \\
			& R4 & 0.348  & 0.622 \\
			& R8 & 0.425  & 0.678  \\
			& R16 & 0.423  & 0.688  \\
			\midrule
			SiamFCv2 & aug & 0.317  & 0.541  \\
			\midrule
			{\footnotesize SiamRPN++} \cite{li2019siamrpn++} & - & 0.461 & 0.634 \\
			{\footnotesize SiamRPN++} & R4 & 0.485 & 0.679  \\
			\midrule
			DiMP18 \cite{bhat2019learning} & - & 0.429 & 0.643\\
			DiMP50 \cite{bhat2019learning} & - & 0.447 & 0.668 \\
			\bottomrule
			
		\end{tabular}
	\end{center}
	\vspace{-1.5em}
	 \caption{Performance scores (success rate `Succ' and precision `Pr' of OPE) for object tracking using different Siamese trackers with regular CNNs as well as RE-SiamNets on Rot-OTB100 dataset. Further, `aug' refers to inclusion of rotation augmentation during training of the tracker model.}
	\label{table_rototb100}
	\vspace{-1.8em}
\end{table} 


\textit{Rot-OTB100. } 
Table \ref{table_rototb100} presents the results for tracking. Adding rotations in the tracked sequences makes tracking considerably harder. Thus, compared to the performance obtained on standard OTB100, the precision and success scores for SiamFC drop by 24.2\% and 26.3\%, respectively. Further, for SiamRPN++, these scores drop by 23.5\% and 28.0\%, respectively. 
Even with just 4 rotational groups RE-SiamNet outperforms both variants of SiamFC comfortably.
Importantly, rotation equivariant Siamese trackers are notably better than standard trackers trained on data with additional rotation augmentations.
Adding rotation equivariance brings improvements even to deep siamese trackers such as SiamRPN++ \cite{li2018high} and yields competitive performance compared to the state-of-the-art DiMP18 and DiMP50 \cite{bhat2019learning}. Additional plots on AUC for precision and success scores are further provided in the supplementary section of the paper.


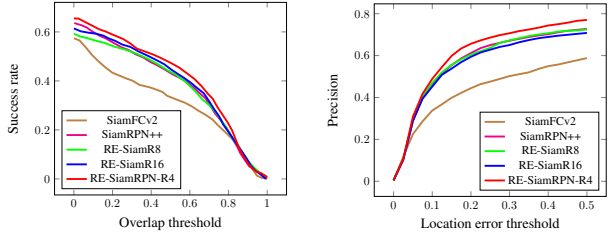
\begin{figure} 
\begin{center}
	\begin{subfigure}{0.49\linewidth}
    \begin{tikzpicture}[scale = 0.45]
        \begin{axis}[%
        		thick,
            legend pos = south west,
            xlabel = {{\large Overlap threshold}},
            ylabel = {{\large Success rate}},
            ]

            \addplot[mark=none,solid, brown, ultra thick] table[x=Xsiamfc, y=Ysiamfc] {data/rob/success.txt};
            \addplot[mark=none,solid, magenta, ultra thick] table[x=Xrpn,y = Yrpn] {data/rob/success.txt};
            \addplot[mark=none,solid, green, ultra thick] table[x=Xresiamr8,y = Ysiamr8] {data/rob/success.txt};
            \addplot[mark=none,solid, blue, ultra thick] table[x=Xresiamr16,y = Ysiamr16] {data/rob/success.txt};
            \addplot[mark=none,solid, red, ultra thick] table[x=Xrpnpp,y = Yrpnpp] {data/rob/success.txt};
       \addlegendentry{SiamFCv2}
       \addlegendentry{SiamRPN++}
       \addlegendentry{RE-SiamR8}
       \addlegendentry{RE-SiamR16}
       \addlegendentry{RE-SiamRPN-R4}
        \end{axis}
    \end{tikzpicture}
   	\label{mto7b}
    \end{subfigure} \hfill
	\begin{subfigure}{0.49\linewidth}
    \begin{tikzpicture}[scale = 0.45]
        \begin{axis}[%
        		thick,
            legend pos = south east,
            xlabel = {{\large Location error threshold}},
            ylabel = {{\large Precision}},
            ]

            \addplot[mark=none,solid, brown, ultra thick] table[x=Index,y = AlexNetV2] {data/rob/precision.txt};
            \addplot[mark=none,solid, magenta, ultra thick] table[x=Index,y = RE-SiamR16] {data/rob/precision.txt};
            \addplot[mark=none,solid, green, ultra thick] table[x=Index,y = RE-SiamR8] {data/rob/precision.txt};
            \addplot[mark=none,solid, blue, ultra thick] table[x=Index,y = RE-SiamRPNpp] {data/rob/precision.txt};
            \addplot[mark=none,solid, red, ultra thick] table[x=Index,y = SiamRPNR4] {data/rob/precision.txt};
       \addlegendentry{SiamFCv2}
       \addlegendentry{SiamRPN++}
       \addlegendentry{RE-SiamR8}
       \addlegendentry{RE-SiamR16}
       \addlegendentry{RE-SiamRPN-R4}
        \end{axis}
    \end{tikzpicture}
   	\label{mto7b}
    \end{subfigure}
\end{center}
\vspace{-2em}
\caption{Performance curves for ROB dataset obtained using SiamFCv2 and RE-SiamNet with different choices of equivariant rotation groups. }
\label{fig_rob_plots}	
\end{figure}

\begin{table}
\small
	\begin{center}
		\begin{tabular}{ l l |c c | c c} 
			\toprule
			 & & \multicolumn{2}{c|}{ROB} &
			 \multicolumn{2}{c}{Rot-OTB100} \\
			Type & Range & SR$_{0.5}$ & SR$_{0.7}$ & SR$_{0.5}$ & SR$_{0.7}$\\ 
			\midrule
			\multirow{3}{*}{Baselines} & $\pm\frac{\pi}{4}$ & 0.25 & 0.25 & 0.25 & 0.25 \\
			& $\pm\frac{\pi}{8}$ & 0.125 & 0.125 & 0.125 & 0.125 \\
			& $\pm\frac{\pi}{16}$ & 0.062 & 0.062 & 0.0625 & 0.062\\
			\midrule
			\multirow{1}{*}{R4} & $\pm\frac{\pi}{4}$ & 0.57 & 0.66 & 0.61 & 0.73 \\
			\midrule
			\multirow{2}{*}{R8} & $\pm\frac{\pi}{8}$ & 0.55 & 0.64 & 0.60 & 0.73  \\
			& $ \pm\frac{\pi}{4}$ & 0.71 & 0.82 & 0.79 & 0.87  \\
			\midrule
			\multirow{3}{*}{R16} & $\pm\frac{\pi}{16}$ & 0.10 &  0.14 & 0.16 & 0.32\\
			& $ \pm\frac{\pi}{8}$ & 0.15 & 0.21 & 0.22 & 0.38  \\	
			& $ \pm\frac{\pi}{4}$ & 0.31 & 0.46 & 0.38 & 0.51  \\
		\bottomrule	
		\end{tabular}
	\end{center}
	\vspace{-1.5em}
	\caption{Performance values for RE-SiamFC with R8 on the task of 2D relative pose estimation for ROT-OTB100 and ROB datasets. Scores reported are in terms of success rate (SR) at IoU thresholds of 0.5 and 0.7. Reported baselines are computed assuming equal probability for each orientation in the dataset.}
	\label{table_pose}
	\vspace{-1.8em}
\end{table}

\textit{ROB.}
We benchmark rotation equivariance also on natural in-plane rotations on the ROB dataset, see Figure \ref{fig_rob_plots}.
It shows the performance plots obtained on ROB dataset using SiamFCv2, SiamRPN++ and their RE-SiamNet equivalents.
We make similar observations as in Rot-OTB100.
Adding rotation equivariance makes both SiamFC and SiamRPN++ more capable to handle natural rotations and overall, the precision and success rates improve.
We provide qualitative examples in \mbox{Figure \ref{fig_rob_results}}, showcasing the benefits of inducing rotation equivariance in Siamese trackers.

\textit{OTB100 and GOT-10k. }To further analyze if the rotation equivariant formulation can have adverse effects on other tracking challenges, we compared the results of RE-SiamFC with 4 rotation groups to that of the base Siamese model on OTB100 and GOT-10k. For both the cases, drops in performance scores were within 2\% of the original values. Such minor drop is expected given that the rotation equivariant trackers use lesser number of channels for the same number of parameters, thereby exhibiting slightly lower discriminative power in general.

\subsection{Unsupervised Pose Estimation}

We experimentally demonstrate that RE-SiamNets can extract the relative 2D pose of the target over time, using the first frame as a reference.
We provide results in Table \ref{table_pose} on the Rot-OTB100 and ROB datasets.
In this experiment, we measure the success rate $SR_\alpha$ as the fraction of frames for which the actual and predicted orientations are within the specified range at an IoU threshold of $\alpha$.

We observe that rotation equivariant trackers recover the relative orientation change with average accuracy above 60\%, well beyond the random baseline.
With 8 rotational groups, RE-SiamNets can even predict angles within a confidence of $\pm \frac{\pi}{8}$ at a similar accuracy.
For finer rotations within $\pm \frac{\pi}{16}$ there is a significant drop, with accuracies ranging between $0.1$ and $0.3$.
The problem is that by increasing the rotation groups, we trade the parameters required for better tracking with parameters that are required for finer rotational bases, thus reducing the final discriminative capacity of our trackers.
We include some qualitative examples in Figure \ref{fig_rob_results} to show the orientations predicted by our rotation equivariant tracker.


\subsection{Rotational-based Motion Constraints}
\begin{table}
\small
	\begin{center}
		\begin{tabular}{ l | l c c  c | c c} 
			\toprule
			& \multicolumn{4}{c|}{Orientation Estimation} & \multicolumn{2}{c}{Tracking} \\ 
			Type & Range & SR$_{0.3}$ & SR$_{0.5}$ & SR$_{0.7}$ & Pr & Succ\\ 
			\midrule
			R8 & \multirow{2}{*}{$\pm\frac{\pi}{4}$}  & 0.72 & 0.79 & 0.87 & 0.42 & 0.68 \\
            c-R8 &  & 0.75 & 0.80 & 0.88 & 0.43 & 0.69\\
			\midrule
			R16 & \multirow{2}{*}{$\pm\frac{\pi}{4}$} & 0.34 & 0.38 & 0.51 & 0.42 & 0.69\\
			c-R16 &  & 0.36 & 0.42 & 0.54 & 0.43 & 0.69 \\
			\bottomrule	
		\end{tabular}
	\end{center}
	\vspace{-1.5em}
	\caption{Accuracy of orientation estimation and performance scores for object tracking on Rot-OTB100 dataset obtained for RE-SiamFC with (denoted with prefix `c-') and without imposing constraint on rotational motion. Here, `Range' refers to permissible change in orientation between two consecutive frames of any video, `SR$_X$' refers to success rate at an IoU threshold of $X$, and `Pr' and `Succ' denote OPE scores for precision and success rate for tracking.}
	\label{table_motion}
	\vspace{-1.5em}
\end{table}

Last, we explore briefly whether the predictions of orientation estimates can be used to improve tracking by an additional constraint to encourage smooth orientation changes over time.
We present results in Table \ref{table_motion}.
Adding the rotation constraint on rotational motion has a modest yet positive influence on tracking performance, while the benefits regarding robustness are higher (data not shown).
We conjecture that introducing other types of equivariance to place more constraints on the attainable types of motion in videos would yield even more robust trackers.


\section{Conclusions}
This paper addresses the challenge of in-plane rotations of the target in visual object tracking.
We demonstrated that frequent in-plane rotations can have an adverse effect on conventional trackers, for which data augmentations do not suffice.
To address this, we introduce rotation equivariant Siamese trackers, specifically for SiamFC and SiamRPN++, that can adapt to rotation changes at no extra parameter cost due to shared weights.
Results show that rotation equivariant Siamese trackers can track accurately under the presense of artificial and natural rotations, they can accurately recognize the relative orientation changes of the target with respect to the first reference frame, and they can even be made more robust by placing additional rotational motion constraints.


\section*{Acknowledgements}
We are thankful to Gabriele Cesa, Maurice Weiler, Daniel Worrall and Arnold Smeulders for the insightful discussions. We further thank Akash Gupta, Silvia Sultana, Rakshita and Akhil for helping with the annotations.

{\small
\bibliographystyle{ieee_fullname}
\bibliography{main_supp}
}

\appendix

\section{Datasets}
We provide here the details related to the three datasets that have been used to benchmark the performance of RE-SiamNets in this paper. The three datasets are Rot-MNIST, Rot-OTB100 and ROB sequences. Details follow below.

\begin{figure*}
    \centering
    \begin{subfigure}{0.19\textwidth}
    \centering
    \includegraphics[width=3.2cm, height=2.4cm]{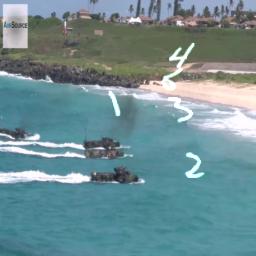}
    \end{subfigure}
    \begin{subfigure}{0.19\textwidth}
    \centering
    \includegraphics[width=3.2cm, height=2.4cm]{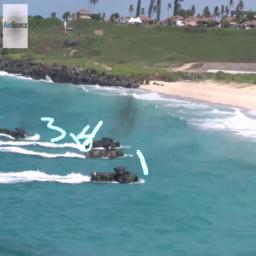}
    \end{subfigure}
    \begin{subfigure}{0.19\textwidth}
    \centering
    \includegraphics[width=3.2cm, height=2.4cm]{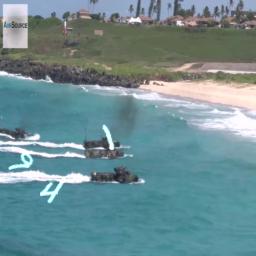}
    \end{subfigure}
    \begin{subfigure}{0.19\textwidth}
    \centering
    \includegraphics[width=3.2cm, height=2.4cm]{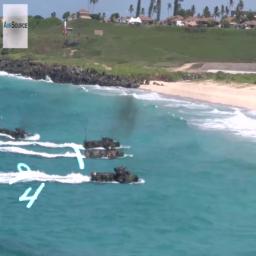}
    \end{subfigure}
    \begin{subfigure}{0.19\textwidth}
    \centering
    \includegraphics[width=3.2cm, height=2.4cm]{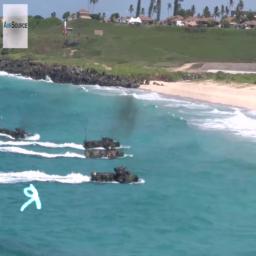}
    \end{subfigure}\\
    \begin{subfigure}{0.19\textwidth}
    \centering
    \includegraphics[width=3.2cm, height=2.4cm]{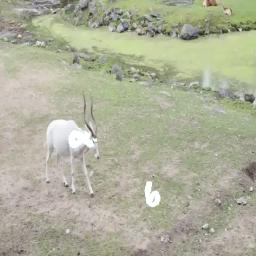}
    \end{subfigure}
    \begin{subfigure}{0.19\textwidth}
    \centering
    \includegraphics[width=3.2cm, height=2.4cm]{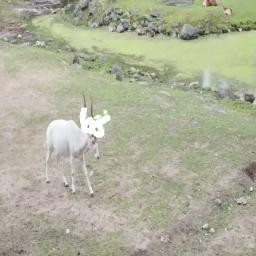}
    \end{subfigure}
    \begin{subfigure}{0.19\textwidth}
    \centering
    \includegraphics[width=3.2cm, height=2.4cm]{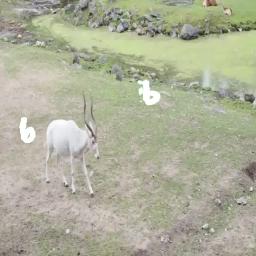}
    \end{subfigure}
    \begin{subfigure}{0.19\textwidth}
    \centering
    \includegraphics[width=3.2cm, height=2.4cm]{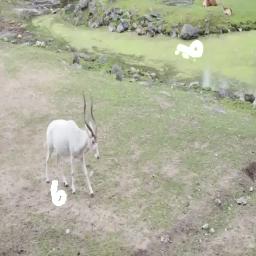}
    \end{subfigure}
    \begin{subfigure}{0.19\textwidth}
    \centering
    \includegraphics[width=3.2cm, height=2.4cm]{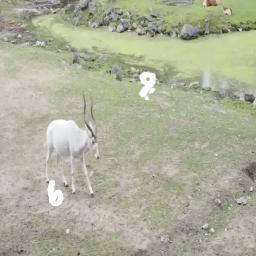}
    \end{subfigure}\\
    \begin{subfigure}{0.19\textwidth}
    \centering
    \includegraphics[width=3.2cm, height=2.4cm]{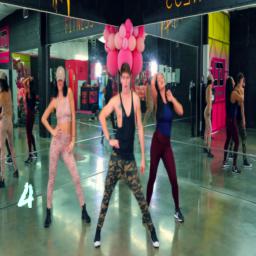}
    \end{subfigure}
    \begin{subfigure}{0.19\textwidth}
    \centering
    \includegraphics[width=3.2cm, height=2.4cm]{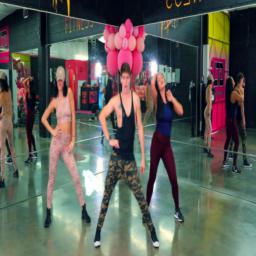}
    \end{subfigure}
    \begin{subfigure}{0.19\textwidth}
    \centering
    \includegraphics[width=3.2cm, height=2.4cm]{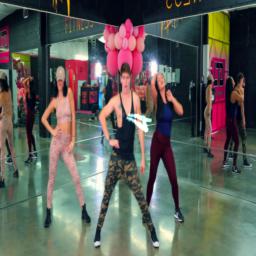}
    \end{subfigure}
    \begin{subfigure}{0.19\textwidth}
    \centering
    \includegraphics[width=3.2cm, height=2.4cm]{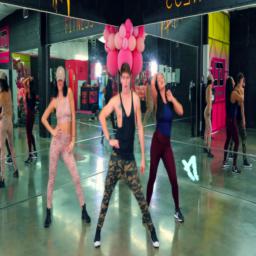}
    \end{subfigure}
    \begin{subfigure}{0.19\textwidth}
    \centering
    \includegraphics[width=3.2cm, height=2.4cm]{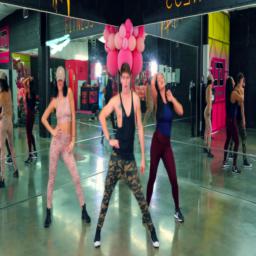}
    \end{subfigure}
    \caption{Sampled frames from 3 sequences of the test set of Rot-MNIST dataset. The backgrounds are taken from sequences of GOT-10k dataset \protect\cite{huang2019got}. Further, to avoid clutter around the target, we have avoided labelling the bounding boxes in the examples above.}
    \label{img_rotmnist_samples}
\end{figure*}
\subsection{Rot-MNIST}
Rot-MNIST, as the name implies, comprises rotating MNIST digits on backgrounds of natural images extracted from the original GOT-10k \cite{huang2019got} training set. Each video comprises MNIST digits floating around and the target digit rotates. Both the motions, translation as well as rotation, are governed by Brownian equation. The training set comprises 2500 sequences exhibiting only translational motion, while the test set contains 100 sequences comprising translation as well as rotation. Each sequence of the training set as well as the test set contains 100 frames. Randomly sampled frames from 3 sequences of the test set are shown in \mbox{Figure \ref{img_rotmnist_samples}}.

\subsection{Rot-OTB100}
Rot-OTB100 is a dataset built through rotating the image frames of the original OTB100 dataset. For this purpose, each sequence is taken and starting from the first frame, every frame is rotated 0.5 degrees counter-clockwise with respect to its previous frame. For keeping the dataset structure similar to that of OTB, we do not use rotated bounding boxes, rather the regular ones. The bounding box is chosen in a way such that the rotated version of the original bounding box (obtained after rotating) fits tightly within it. Thus, Rot-OTB100 is a testset with exactly same number of sequences as OTB100, except that the frames are rotated.

\subsection{ROB}
Details of Rotating Object Benchmark (RTB) as well as example sequences have already been shown in the main paper. All the 35 videos have been acquired at 10 fps and comprise 300-500 frames each. 

\section{Implementation details}
\subsection{Models}

The discussion on the model details are provided below with respect to the three benchmarking datasets as outlined earlier. 

\textbf{Rot-MNIST. } For Rot-MNIST, we develop a reduced rotation equivariant variant of SiamFC \cite{bertinetto2016fully}, comprising 999K paramters. RE-mSiamFC differs in terms of the size of the kernels used. Comparisons are made with the non-rotated equivariant of the same model comprising equal number of parameters. For 4 rotational groups (R4), the resultant model comprises 5 convolutional layers, comprising 62, 75, 157 and 160, and kernels of sizes $3 \times 3$ in all the layers. Note that the number of channels for any choice of rotational groups is made such that the number of parameters is approximately 999K. Further, all except the last layer are followed by Batchnorm and ReLU activation layers. Finally, pooling is used across the rotational groups to obtain a single set of feature maps from the last layer.

The baseline model for comparison is the non-rotational equivariant version comprising similar number of parameters. This model is referred as mSiamFC. The model is similar to that of RE-mSiamFC, except two differences. First, all rotation equivariant modules are replaced with non-rotation equivariant counterparts. Further, the 4 layers comprise 96, 128, 256 and 256 in the four layers, respectively.

\textbf{Rot-OTB100 and ROB. }Compared to Rot-MNIST, Rot-OTB100 and ROB datasets are relatively more complex and larger models are needed. In this regard, we build two rotation equivariants of SiamFC, referred as \emph{RE-SiamFC} and {RE-SiamFCv2}. RE-SiamFC has an architecture similar to that of the original SiamFC \cite{bertinetto2016fully}, and comprises 2.33M parameters approximately. The 5 convolutional layers comprise 72, 160, 240, 240 and 160 channels, respectively. The respective kernel sizes are $11 \times 11$ and $5 \times 5$ in the first two layers, and $3 \times 3$ in the last three layers. The choice of padding, stride and pooling is similar to that of SiamFC \cite{bertinetto2016fully}, but rotation equivariant. The number of channels stated above are for R4. For R8 and R16, these are scaled down keeping the number of parameters same. 

RE-SiamFCv2 is similar to RE-SiamFC, except that it uses larger kernel sizes, thus reduced number of channels, thereby keeping the number of parameters equal to 2.33M. It uses 4 convolutional layers, and for R4, these layers are composed of 64, 96, 128 and 163 channels, respectively. The corresponding kernel sizes are $9 \times 9$, $7 \times 7$, $7 \times 7$ and $6 \times 6$, respectively. Accordingly, the number of channels for R8 and R16 are 49, 71, 85, 118 and 36, 48, 60, 80, respectively. All except the last convolutional batchnorm and ReLU activation layers, and pooling is performed across the different groups after the last convolutional layer. 

\subsection{Extension: Training details. }To train RE-mSiamFC, RE-SiamFCv1 and RE-SiamFCv2, we follow a training procedure similar to the default SiamFC \cite{bertinetto2016fully}\footnote{For SiamFC, we use the Pytorch code available at \texttt{https://github.com/huanglianghua/siamfc-pytorch}.} Each model was trained for 50 epochs with batch size of 8 on a single NVIDIA GTX GPU. For R16 variants, we use batch sizes of 8 and 150 epochs. The initial learing rate is set to 1e-2 and it is decayed to 1e-5 during the course of training. The weight decay and momentum terms are set to 1e-4 and 0.9, respectively. For training the models, we use GOT-10k training set.

For RE-SiamRPN++, we follow training details similar to the baseline SiamRPN++ model. We separately trained a ResNet50 architecture using rotation equivariant modules. This backbone was trained for 50 epochs using batch sizes of 128. The details of model training are same as the standard training of SiamRPN++, as specified in \texttt{pysot}\footnote{For SiamRPN++ and its rotation equivariant modifications, we use \texttt{pysot} library available at \texttt{https://github.com/STVIR/pysot}.} pytorch library.

\section{Extension: Results}
\begin{figure*}
    \centering
    \begin{subfigure}{0.24\textwidth}
    \centering
    \includegraphics[width=4cm, height=2.4cm]{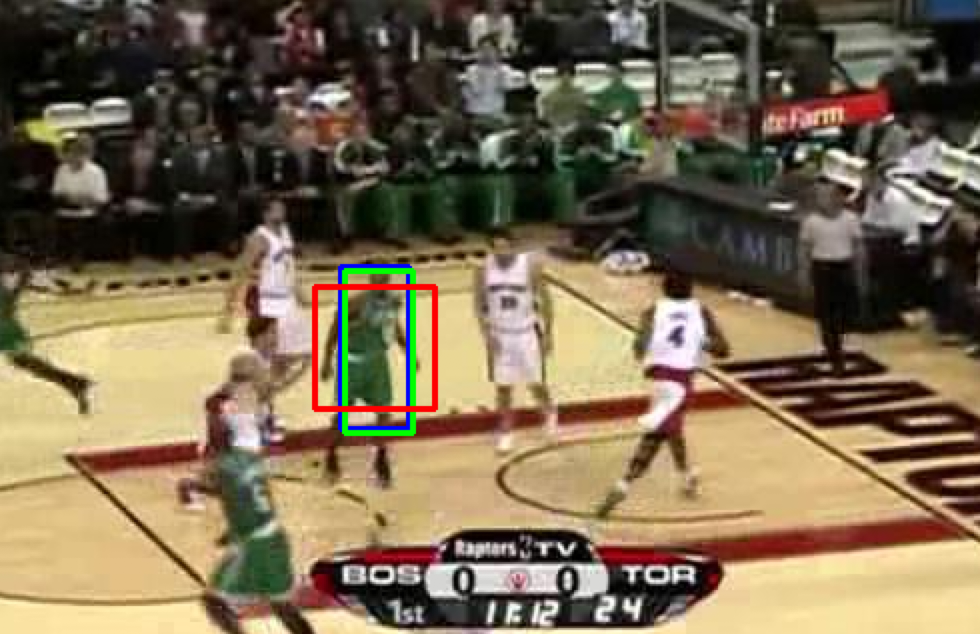}
    \end{subfigure}
    \begin{subfigure}{0.24\textwidth}
    \centering
    \includegraphics[width=4cm, height=2.4cm]{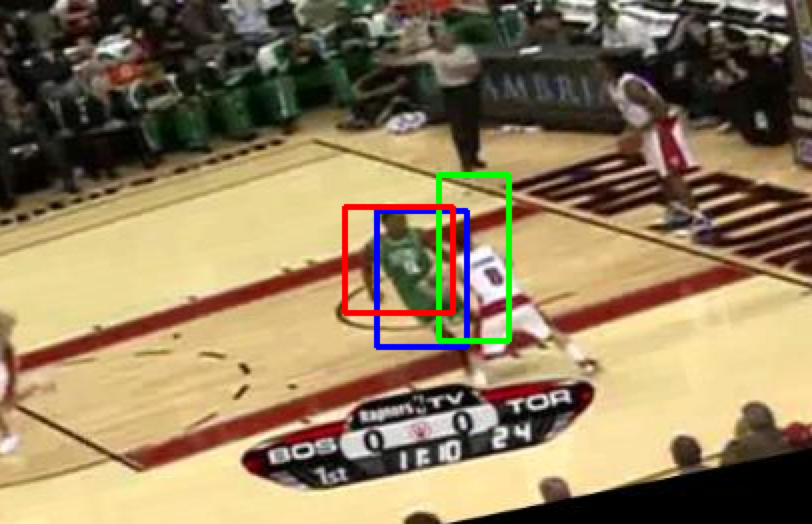}
    \end{subfigure}
    \begin{subfigure}{0.24\textwidth}
    \centering
    \includegraphics[width=4cm, height=2.4cm]{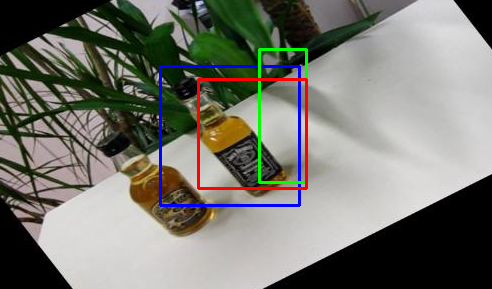}
    \end{subfigure}
    \begin{subfigure}{0.24\textwidth}
    \centering
    \includegraphics[width=4cm, height=2.4cm]{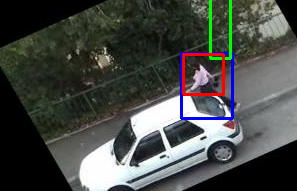}
    \end{subfigure}
    \begin{subfigure}{0.24\textwidth}
    \centering
    \includegraphics[width=4cm, height=2.4cm]{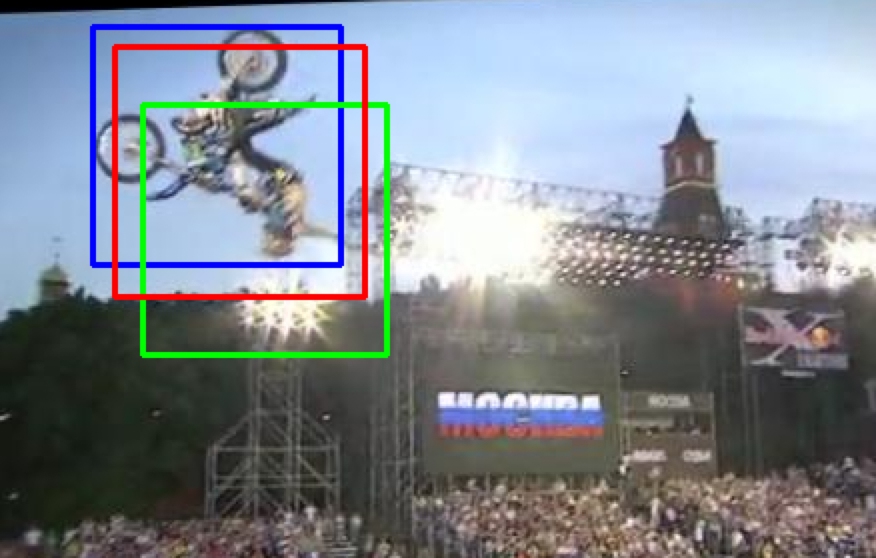}
    \end{subfigure}
    \begin{subfigure}{0.24\textwidth}
    \centering
    \includegraphics[width=4cm, height=2.4cm]{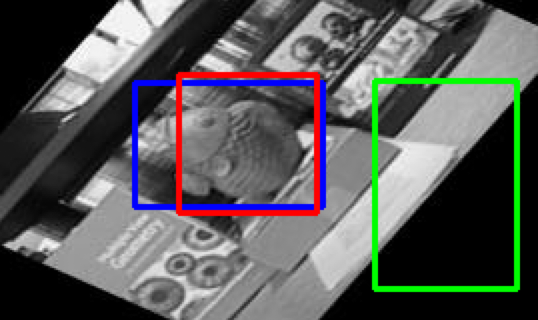}
    \end{subfigure}
    \begin{subfigure}{0.24\textwidth}
    \centering
    \includegraphics[width=4cm, height=2.4cm]{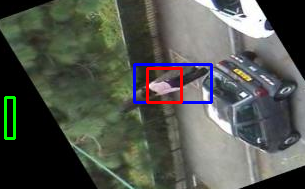}
    \end{subfigure}
    \begin{subfigure}{0.24\textwidth}
    \centering
    \includegraphics[width=4cm, height=2.4cm]{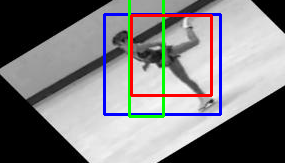}
    \end{subfigure}
    \caption{Results on Rot-OTB100 obtained with SiamFC-Netv2 (green) and RE-SiamFCv2 with R8 (red).}
    \label{results_rototb_resiamfc}
\end{figure*}

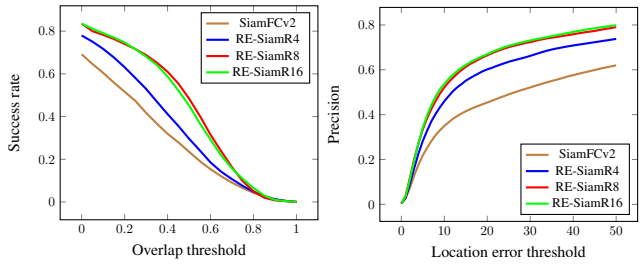
\begin{figure} 
\begin{center}
	\begin{subfigure}{0.49\linewidth}
    \begin{tikzpicture}[scale = 0.5]
        \begin{axis}[%
        		thick,
            legend pos = north east,
            xlabel = {{\large Overlap threshold}},
            ylabel = {{\large Success rate}},
            ]

            \addplot[mark=none,solid, brown, ultra thick] table[x=Threshold,y = AlexNetV2] {data/rot-otb/resiamfc_success.txt};
            \addplot[mark=none,solid, blue, ultra thick] table[x=Threshold,y = GCNN-N4] {data/rot-otb/resiamfc_success.txt};
            \addplot[mark=none,solid, red, ultra thick] table[x=Threshold,y = GCNN-N8] {data/rot-otb/resiamfc_success.txt};
            \addplot[mark=none,solid, green, ultra thick] table[x=Threshold,y = GCNN-N16] {data/rot-otb/resiamfc_success.txt};
       \addlegendentry{SiamFCv2}
       \addlegendentry{RE-SiamR4}
       \addlegendentry{RE-SiamR8}
       \addlegendentry{RE-SiamR16}
        \end{axis}
    \end{tikzpicture}
   	\label{mto7b}
    \end{subfigure} \hfill
	\begin{subfigure}{0.49\linewidth}
    \begin{tikzpicture}[scale = 0.5]
        \begin{axis}[%
        		thick,
            legend pos = south east,
            xlabel = {{\large Location error threshold}},
            ylabel = {{\large Precision}},
            ]

            \addplot[mark=none,solid, brown, ultra thick] table[x=Threshold,y = AlexNetV2] {data/rot-otb/resiamfc_precision.txt};
            \addplot[mark=none,solid, blue, ultra thick] table[x=Threshold,y = GCNN-N4] {data/rot-otb/resiamfc_precision.txt};
            \addplot[mark=none,solid, red, ultra thick] table[x=Threshold,y = GCNN-N8] {data/rot-otb/resiamfc_precision.txt};
            \addplot[mark=none,solid, green, ultra thick] table[x=Threshold,y = GCNN-N16] {data/rot-otb/resiamfc_precision.txt};
       \addlegendentry{SiamFCv2}
       \addlegendentry{RE-SiamR4}
       \addlegendentry{RE-SiamR8}
       \addlegendentry{RE-SiamR16}
        \end{axis}
    \end{tikzpicture}
   	\label{mto7b}
    \end{subfigure}
\end{center}
\vspace{-3em}
\caption{Performance curves for Rot-OTB100 dataset obtained using SiamFCv2 and RE-SiamNet with different choices of equivariant rotation groups. All networks chosen here used 233K optimization parameters. }
\label{fig_rototb_plots}	
\end{figure}

We show here a few additional results related to our experiments. Figure \ref{results_rototb_resiamfc} shows a few examples of predictions made by SiamFCv2 as well as the equivalent RE-SiamFCv2 variant. Further, we provide precision and success plots of OPE for SiamFCv2-R8 on Rot-OTB100 in Figure \ref{fig_rototb_plots}. Further, we show success rates at different overlap thresholds on different orientation estimates in Table \ref{table_rototb100}.  

\begin{table}
\small
	\begin{center}
		\begin{tabular}{ l l |c c c  c c} 
			\toprule
			Type & Range & SR$_{0.1}$ & SR$_{0.3}$ & SR$_{0.5}$ & SR$_{0.7}$ & SR$_{0.9}$\\ 
			\midrule
			\multirow{1}{*}{R4} & $\pm\frac{\pi}{4}$ & 0.52 & 0.56 & 0.61 & 0.73 & 0.95\\
			\midrule
			\multirow{2}{*}{R8} & $\pm\frac{\pi}{8}$ & 0.48 & 0.52 & 0.60 & 0.73 & 0.95 \\
			& $ \pm\frac{\pi}{4}$ & 0.67 & 0.72 & 0.79 & 0.87 & 0.98  \\
			\midrule
			\multirow{3}{*}{R16} & $\pm\frac{\pi}{16}$ & 0.12 &  0.14 & 0.16 & 0.32 & 0.87\\
			& $ \pm\frac{\pi}{8}$ & 0.17 & 0.20 & 0.22 & 0.38 & 0.88 \\	
			& $ \pm\frac{\pi}{4}$ & 0.30 & 0.34 & 0.38 & 0.51 & 0.92  \\
		\bottomrule	
		\end{tabular}
	\end{center}
	\caption{Performance scores measured in terms of success rate at different overlap thresholds for orientation estimation using SiamFCv2 for Rot-OTB100.}
	\label{table_rototb100}
\end{table}

\end{document}